\documentclass[conference]{IEEEtran}
\IEEEoverridecommandlockouts
\usepackage{cite}
\usepackage{amsmath,amssymb,amsfonts}
\usepackage{algorithm}
\usepackage{algpseudocode}
\usepackage{graphicx}
\usepackage{textcomp}
\usepackage{xcolor}
\usepackage[a4paper, total={184mm,239mm}]{geometry}
\def\BibTeX{{\rm B\kern-.05em{\sc i\kern-.025em b}\kern-.08em
    T\kern-.1667em\lower.7ex\hbox{E}\kern-.125emX}}
\begin{document}

\title{Fluid Dynamic DNNs for Reliable and Adaptive Distributed Inference on Edge Devices \\
}

\author{

\IEEEauthorblockN{Lei Xun*\IEEEauthorrefmark{2}, Mingyu Hu*\IEEEauthorrefmark{2}, Hengrui Zhao\IEEEauthorrefmark{2},  Amit Kumar Singh\IEEEauthorrefmark{3}, Jonathon Hare\IEEEauthorrefmark{2}, Geoff V. Merrett\IEEEauthorrefmark{2}}
\IEEEauthorblockA{\IEEEauthorrefmark{2}\textit{School of Electronics and Computer Science, University of Southampton, UK} \\
\IEEEauthorrefmark{3}\textit{School of Computer Science and Electronic Engineering, University of Essex, UK}\\
Email: \{l.xun, mh1u20, hz20u22\}@soton.ac.uk, a.k.singh@essex.ac.uk, \{jsh2, gvm\}@ecs.soton.ac.uk}
* Equal Contributions
\vspace{-5mm}
}
\maketitle

\begin{abstract}
Distributed inference is a popular approach for efficient DNN inference at the edge. However, traditional Static and Dynamic DNNs are not distribution-friendly, causing system reliability and adaptability issues. In this paper, we introduce Fluid Dynamic DNNs (Fluid DyDNNs), tailored for distributed inference. Distinct from Static and Dynamic DNNs, Fluid DyDNNs utilize a novel nested incremental training algorithm to enable independent and combined operation of its sub-networks, enhancing system reliability and adaptability. Evaluation on embedded Arm CPUs with a DNN model and the MNIST dataset, shows that in scenarios of single device failure, Fluid DyDNNs ensure continued inference, whereas Static and Dynamic DNNs fail. When devices are fully operational, Fluid DyDNNs can operate in either a High-Accuracy mode and achieve comparable accuracy with Static DNNs, or in a High-Throughput mode and achieve 2.5x and 2x throughput compared with Static and Dynamic DNNs, respectively.
\end{abstract}

\begin{IEEEkeywords}
Fluid Dynamic DNNs, Distributed Inference
\vspace{-2mm}
\end{IEEEkeywords}

\section{Introduction}
\vspace{-1mm}
DNN inference is increasingly being executed on edge devices due to advantages of latency, privacy and always-on availability. However, DNNs are computationally intensive, which makes efficient edge deployments a challenge \cite{xun2020optimising}. Previous works have addressed this challenge through static model compression to fit DNNs into available hardware resources \cite{yang2018netadapt}, and Dynamic DNNs \cite{xun2019incremental, yu2019universally, dynamic-ofa, bouzidi2023hadas} (Fig. \ref{fig: Overview of Fluid DNN}a) which contain multiple switchable sub-networks with different widths and latency/accuracy trade-offs. They can adapt to dynamically changing available hardware resources at runtime. 

Distributed inference is an orthogonal approach, which partitions a DNN into sub-tasks and deploys them on multiple devices for parallel computing. Existing works mostly follow a distributed framework in which one device is selected as the \textbf{Master} for making decisions of partitioning and distribution, while other devices used as \textbf{Worker} for receiving and executing the workloads \cite{mao2017modnn, zeng2020coedge}. However, existing Static and Dynamic DNNs (Fig. \ref{fig: Overview of Fluid DNN}a) are not distribution-friendly in this framework due to the high dependencies between their sub-networks to get correct inference, leading to low adaptability on device usage.

Meanwhile, physical devices in a distributed system could completely fail due to factors such as power outages and hardware/software failures, leading to reliability issues \cite{yousefpour2020resilinet}. When the Worker fails (Fig. \ref{fig: Overview of Fluid DNN}b), the Static DNNs on the Master also fail because it doesn't have enough resources on its own. In contrast, the Dynamic DNNs can adapt by switching to a smaller sub-network (e.g. 50\% model) on the Master. This allows it to keep the expected performance with a temporary accuracy loss. However, when the Master fails (Fig. \ref{fig: Overview of Fluid DNN}c), the Static and Dynamic DNNs on the Worker fail too because these model weights cannot be used independently. To further address the system reliability and adaptability issues, we introduce Fluid DyDNNs, and the main contributions of this paper are:

\begin{figure}[t]
\centerline{\includegraphics[width=1\columnwidth]{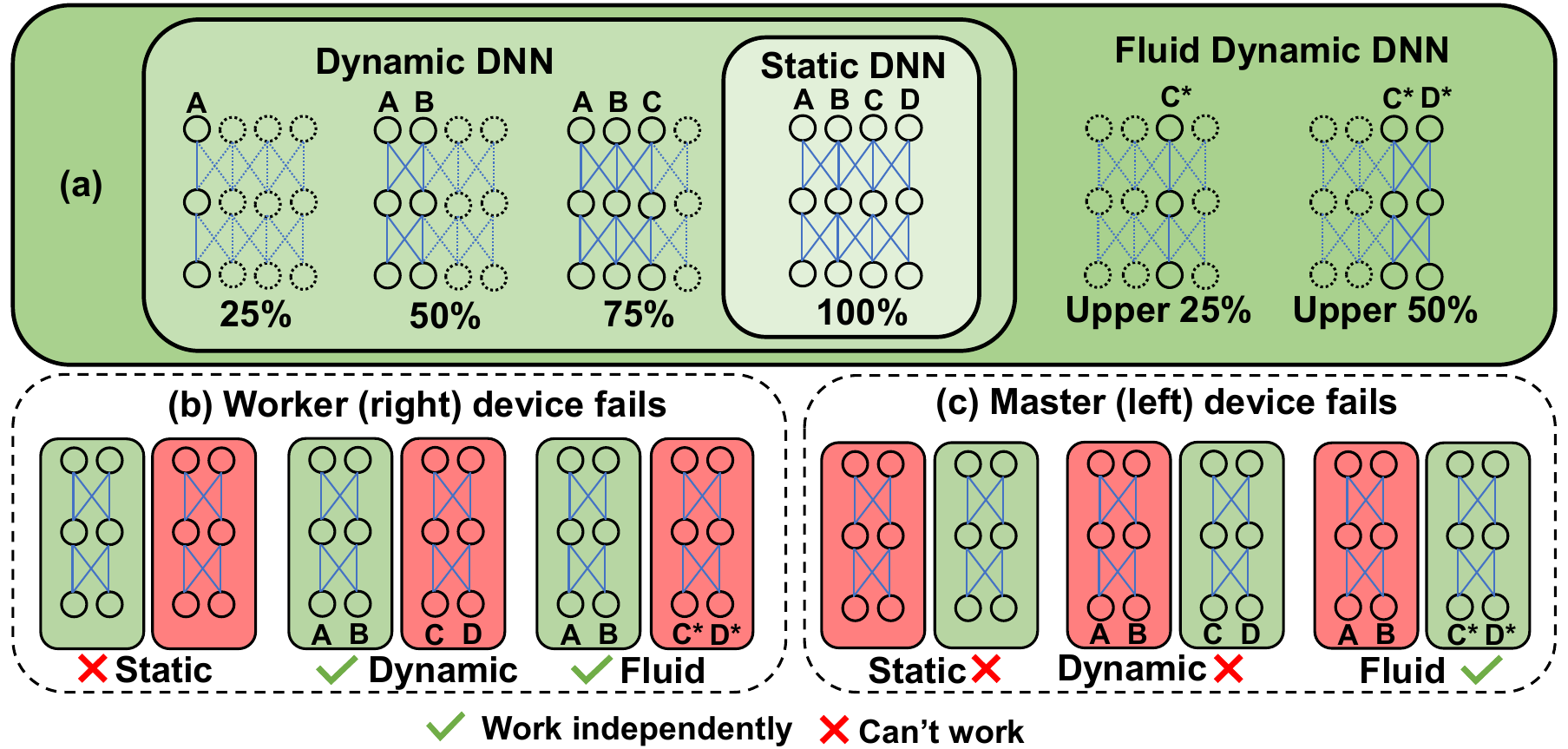}}
\vspace{-3mm}
\caption{(a) Overview of three types of DNN model (b) under worker device failure and (c) master device failure scenarios of distributed DNN inference.}
\vspace{-6mm}
\label{fig: Overview of Fluid DNN}
\end{figure}

\begin{itemize}
    \item A Fluid DyDNN model trained by a novel nested incremental training algorithm, reducing dependencies between sub-networks and enhancing reliability and adaptability.
    \item Experimental results show it can maintain inference during single-device failures. With both devices online, it can adapt to either match Static DNNs' accuracy or improve throughput by up to 2.5x with temporary accuracy loss.
\end{itemize}

\section{Fluid Dynamic Neural Networks}
\vspace{-1mm}
Fluid DyDNNs (Fig. \ref{fig: Overview of Fluid DNN}a) reduce the dependencies between their sub-networks. Unlike traditional Dynamic DNNs, where larger sub-networks contain and depend on the weights of smaller ones, Fluid DyDNNs operate with a modular design. This design has discrete sub-networks, in an example of 4 sub-networks, the additional upper 25\% and 50\% model derive their weights from the 50-75\% and 50-100\% model, respectively. These sub-networks can function autonomously or be integrated with the lower 50\% model to enhance accuracy, forming more accurate 75\% and 100\% models.

\begin{algorithm}
\caption{Nested Incremental Training}
\begin{algorithmic}[1]
\For{$i \gets 1, n_{iters} $}
\For{model in \textit{25\%, 50\%, 75\%, 100\% models}}
   \State Train the model
       \State Copy trained weights to the next model 
\EndFor
\For{model in upper 25\% and upper 50\% models}
   \State Copy corresponding weights from 100\% model
   \State Re-train the model
   \State Copy the re-trained weights back to 100\% model
\EndFor
\EndFor
\end{algorithmic}
\end{algorithm}

\subsection{Nested Incremental Training}
\vspace{-1mm}
To train a Fluid DyDNN model, we introduce a novel nested incremental training method outlined in Algorithm 1, with an example of 4 sub-networks but it is applicable to any number. Initially,  A Dynamic DNN with 4 sub-networks (i.e. [25\%, 50\%, 75\%, 100\%] models) is first trained incrementally as the base model (line 2-5), then a nested Dynamic DNN (upper 25\% and upper 50\% models) is trained incrementally so they can be used independently (line 6-10). Reusing the weights from the upper 25\%/50\% models on the 75\%/100\% models is nontrivial; therefore, we fine-tune all the models for multiple iterations.

\subsection{Reliable and Adaptive Distributed Inference}
\vspace{-1mm}
The trained sub-networks can either work independently (i.e. [25\%, 50\%] and [upper 25\%, upper 50\%] models) or be combined to inference collectively (i.e. [75\%, 100\%] models). In independent mode, devices receive separate inputs in parallel, maximizing system throughput (High-Throughput (HT) mode). In collective mode, devices process the same inputs jointly, ensuring peak accuracy (High-Accuracy (HA) mode). Fluid DyDNNs seamlessly transition between two modes to meet varying performance demands and adapt to available resources.

\section{Experimental Results and Analysis}
\begin{figure}[t]
\centerline{\includegraphics[scale=0.39]{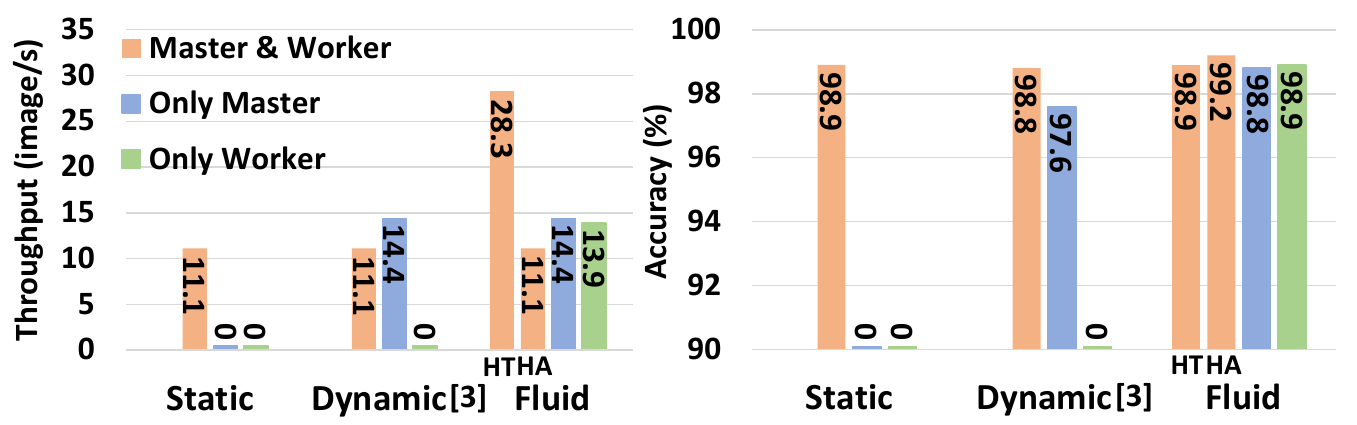}}
\vspace{-3mm}
\caption{Experimental results of throughput and accuracy of Static DNNs, Dynamic DNNs \cite{xun2019incremental} and Fluid DyDNNs under High-Accuracy (HA) and High-Throughput (HT) mode. Inference results were collected when only the Master is online, only the Worker is online, and when both devices are online.}
\label{fig: Throughput and Accuracy}
\vspace{-5mm}
\end{figure}

\vspace{-1mm}
The proposed Fluid DyDNNs were validated on a small DNN model with three convolution layers and one fully-connected layer. The [25\%, 50\%, 75\%, 100\%] sub-networks have [4,8,12,16] (3x3) kernels, respectively. They were trained using Algorithm 1 and tested on the MNIST dataset. We also trained a Static DNN and a Dynamic DNN (using incremental training \cite{xun2019incremental}. The throughput was measured by running the models on the CPU of a Nvidia Jetson Xavier NX platform. For distributed inference, we used TCP to achieve data exchange. To simplify the runtime scenario and avoid network variance, we measured the communication latency offline. The total throughput of the system can be calculated with the sum of computation and communication latency. Fig. \ref{fig: Throughput and Accuracy} shows the throughput and accuracy of Static, Dynamic DNNs and Fluid DyDNNs under two execution scenarios:

\textbf{Single-Device Failures:} Static DNNs exhibit the lowest reliability; failure of either the Master or Worker results in system failure, because the remaining device's model weights cannot function independently, shown as zero throughput and accuracy in Fig. \ref{fig: Throughput and Accuracy}. Dynamic DNNs offer improved reliability; the Master's sub-networks can continue to inference with temporarily reduced accuracy (due to the loss of upper 50\% model weights) when the Worker fails. However, if the Master fails, the system cannot operate since the upper 50\% model requires the lower 50\% model weights (on the Master) to function. In contrast, Fluid DyDNNs show superior reliability; their independent sub-networks allow for continuous inference when any one device fails. The temporary accuracy losses when using smaller sub-networks are recoverable whenever the system can re-deploy larger sub-networks.

\textbf{No Devices Failure:} when both devices are operational, Static DNNs are limited to a throughput of 11.1 image/s (Fig. \ref{fig: Throughput and Accuracy}) due to inevitable communication overhead between devices. Dynamic DNNs can adapt to a single device by deploying smaller sub-networks (e.g. 50\% model on the Master), sacrificing accuracy temporarily for increased throughput, up to 14.4 image/s. Fluid DyDNNs showcase maximum adaptability; in HT mode, they operate two independent sub-networks in parallel across both devices, reaching 28.3 image/s, this is 2.5x the throughput of Static DNNs and 2x that of Dynamic DNNs. In HA mode, they can replicate the distributed Static DNNs to recover to its peak accuracy. Fluid DyDNNs outperform Static and Dynamic DNNs in peak accuracy, likely due to the training regularization benefits from their additional sub-networks.

\section{Conclusions}
\vspace{-1mm}
This work introduces Fluid DyDNNs, a novel approach for distributed inference on edge devices. Our findings indicate they can enhance system reliability in single-device failure scenarios. With both devices online, it can adapt to either high accuracy or boost throughput with temporary accuracy loss.

\section{Acknowledgments}
\vspace{-1mm}
This work is supported by the Engineering and Physical Sciences Research Council (EPSRC) under Grant EP/S030069/1. Experimental data can be found at: https://doi.org/10.5258/SOTON/D2886.


\end{document}